# Double-Constraint Diffusion Model with Nuclear Regularization for Ultra-low-dose PET Reconstruction

Mengxiao Geng, Ran Hong, Bingxuan Li, Qiegen Liu, *Senior Member, IEEE*

*Abstract*—Ultra-low-dose positron emission tomography (PET) reconstruction holds significant potential for reducing patient radiation exposure and shortening examination times. However, it may also lead to increased noise and reduced imaging detail, which could decrease the image quality. In this study, we present a Double-Constraint Diffusion Model (DCDM), which freezes the weights of a pre-trained diffusion model and injects a trainable double-constraint controller into the encoding architecture, greatly reducing the number of trainable parameters for ultra-low-dose PET reconstruction. Unlike full fine-tuning models, DCDM can adapt to different dose levels without retraining all model parameters, thereby improving reconstruction flexibility. Specifically, the two constraint modules, named the Nuclear Transformer Constraint (NTC) and the Encoding Nexus Constraint (ENC), serve to refine the pre-trained diffusion model. The NTC leverages the nuclear norm as an approximation for matrix rank minimization, integrates the low-rank property into the Transformer architecture, and enables efficient information extraction from low-dose images and conversion into compressed feature representations in the latent space. Subsequently, the ENC utilizes these compressed feature representations to encode and control the pre-trained diffusion model, ultimately obtaining reconstructed PET images in the pixel space. In clinical reconstruction, the compressed feature representations from NTC help select the most suitable ENC for efficient unknown low-dose PET reconstruction. Experiments conducted on the *UDPET* public dataset and the *Clinical* dataset demonstrated that DCDM outperforms state-of-the-art methods on known dose reduction factors (DRF) and generalizes well to unknown DRF scenarios, proving valuable even at ultra-low dose levels, such as 1% of the full dose.

## I. INTRODUCTION

Positron emission tomography (PET) is a crucial imaging modality widely used for tumor detection and neurological disorder diagnosis [1], [2]. This technique employs radiolabeled tracers such as $^{18}$F-Fluorodeoxyglucose ($^{18}$F-FDG), which become metabolically incorporated into tissues and emit positrons during decay. The resulting annihilation photons are detected to generate images reflecting regional metabolic activity [3]. While reducing radiotracer dosage is important for minimizing patient radiation exposure, this optimization often compromises image quality. Ultra-low-dose PET, which corresponds to a dose reduction factor (DRF) of greater than 50 (i.e., below 2% of the full dose), exacerbates this trade-off. This results in images with reduced signal-to-noise ratios, increased artifacts, and loss of fine structural details. Moreover, whole-body imaging, which is more structurally complex and physiologically heterogeneous than regional scans, poses greater challenges for accurate reconstruction under extreme dosage constraints. Consequently, maintaining image quality in whole-body ultra-low-dose PET reconstruction stands as a critical unmet challenge in medical imaging research [4].

Traditional reconstruction methods, such as Gaussian filtering [5], total variation (TV) [6], [7], non-local means (NLM) [8], [9], and block-matching and 3D filtering (BM3D) [10], face challenges in producing high-quality images from ultra-low-dose PET acquisitions due to the significant noise amplification associated with low-count data. Recent advancements in deep learning (DL) techniques have shown promise in addressing this challenge [11], [12]. For example, Xu *et al*. [13] implemented a U-Net architecture [14] to reconstruct low-dose PET images. Peng *et al*. [15] developed a 3D U-Net-based method that incorporated CT images as an input to enhance image quality. Chen *et al*. [16] proposed a 3D image space shuffle U-Net by introducing the shuffle/unshuffled layers into the U-Net architecture for ultra-low-dose reconstruction. Furthermore, Li *et al*. [4] replaced convolutional neural networks (CNNs) with vision Transformer (ViT) [17] based on the U-Net framework to learn end-to-end mapping relationships. Gong *et al*. [18] introduced a deep neural network for PET image denoising using simulation data and fine-tuning the last few layers with real datasets. Ouyang *et al*. [19] utilized vanilla generative adversarial network (GAN) with task-specific perceptual loss for ultra-low-dose PET reconstruction via adversarial learning [20]-[22] and integrated a pre-trained Amyloid status classifier for further guidance. Wang *et al*. [23] introduced a 3D multi-modality edge-aware Transformer-GAN from low-quality PET and T1-weighted MR images, reducing radiation risk while improving image quality. Cui *et al*. [24] proposed a prior knowledge-guided triple-domain Transformer-GAN to directly reconstruct high-quality images, integrating knowledge from the sinogram, image, and frequency domains. Overall, these methods learn the one-to-one mapping between low-dose and full-dose images through loss function optimization, often relying on paired datasets for end-to-end training. However, such approaches often

This work was supported in part by the National Key Research and Development Program of China under Grants 2023YFF1204300 and 2023YFF1204302, the National Natural Science Foundation of China under Grant 62122033, and the Key Research and Development Program of Jiangxi Province under Grant 20212BBE53001. (M. Geng and R. Hong are co-first authors.) (Corresponding authors: B. Li and Q. Liu)

M. Geng, R. Hong, and Q. Liu are with School of Information Engineering, Nanchang University, Nanchang 330031, China. (mxiaogeng@163.com, ranhong@email.ncu.edu.cn, liuqiegen@ncu.edu.cn)

B. Li is with Anhui Province Key Laboratory of Biomedical Imaging and Intelligent Processing, Institute of Artificial Intelligence, Hefei Comprehensive National Science Center, Hefei 230088, China. (libingxuan@iai.ustc.edu.cn)

rely on large amounts of paired training data and may fail to generalize well to unknown DRF cases. Furthermore, ultra-low-dose PET reconstruction presents greater challenges than low-dose reconstruction, primarily because the degraded images contain far less usable texture and structural information. This reduction in available information widens the gap between low-quality and high-quality images, making it difficult to achieve accurate results using full-dose images for supervision.

Diffusion models, a class of deep generative models, have recently emerged as a powerful tool for PET reconstruction [25]-[27]. Unlike CNNs and GANs, diffusion models do not require paired training data and can effectively learn complex data distributions through an iterative reverse denoising process [28]-[30]. For example, Jiang *et al.* [31] designed an unsupervised PET enhancement framework based on the latent diffusion model, trained solely on full-dose PET images and enhanced by PET image compression, Poisson diffusion replacement of Gaussian diffusion, and CT-guided cross-attention. Han *et al.* [32] presented a diffusion probabilistic model-based PET reconstruction framework with a coarse prediction module and an iterative refinement module, and they boosted the model with auxiliary guidance and contrastive diffusion strategies. At the same time, a series of studies have applied the diffusion model to the multi-modality reconstruction of PET images. Xie *et al.* [33]-[35] introduced a joint diffusion attention model that combines high-field and ultra-high-field MR images to generate PET images, employing joint probability distribution and attention techniques. Moreover, Pan *et al.* [36] proposed a diffusion-based PET consistency model that improves low-dose PET image quality by learning a consistency function in reverse diffusion and using shifted windows as visual transformers. Xie *et al.* [37] developed a dose-aware diffusion model for 3D low-dose PET imaging by using neighboring slices as conditional information. Diffusion models provide a way for ultra-low-dose PET reconstruction by addressing the challenge of acquiring paired low- dose and full-dose PET data and enabling the use of prior information to narrow the gap. Nevertheless, the high levels of noise and artifacts in ultra-low-dose PET images mean that simply using them to constrain and guide the generation process can introduce interferences, leading to generation errors.

In this study, we propose a **D**ouble-**C**onstraint **D**iffusion **M**odel (DCDM) for achieving high-quality PET imaging while significantly reducing radiation exposure risks. Specifically, the two constraint modules include a nuclear Transformer constraint (NTC) and an encoding nexus constraint (ENC). By integrating the trainable double-constraint controller into the encoding architecture of a pre-trained diffusion model, DCDM reduces the number of trainable parameters while enhancing reconstruction flexibility compared to full fine-tuning models. Additionally, the proposed method can reconstruct low-dose PET images with unknown DRF by leveraging the feature extraction and classification capabilities of the NTC module. The main contributions of this work are summarized as follows:

● *Efficiency of NTC Module for Sparsity and Low-rank Feature Extraction.* The NTC module is designed with a self-attention mechanism and nuclear regularization, which respectively captures long-range dependencies and maintains the sparsity and low-rank properties of feature representations. Compared with the traditional vision Transformer, NTC demonstrates more efficient feature redundancy reduction and better feature extraction performance.

● *NTC-ENC Dual-Constraint Collaboration for Pixel-Latent Space.* A double-constraint controller is designed to regulate the pre-trained diffusion model using prior information for various dose levels, enhancing flexibility and reducing training costs. Specifically, NTC extracts sparse and low-rank feature representations in the latent space. Meanwhile, ENC receives these representations, integrates structural spatial information, and injects them into the diffusion model in the pixel space.

● *Adaptive Reconstruction Framework for an Unknown DRF Scenario.* A dedicated reconstruction framework is proposed to address the challenge of reconstructed images with unknown DRF in clinical practice. The compressed features generated by NTC not only helps ENC with precise control but also accurately classifies input images, adaptively selecting the suitable ENC for PET reconstruction at unknown DRF.

The remainder of this paper is exhibited as follows: Section II introduces some relevant works in this study. Section III contains the key idea of the proposed method. Experimental settings and results are shown in Section IV. Section V conducts a concise discussion, and Section VI draws a conclusion.

## II. PRELIMINARY

### A. Diffusion Models

Diffusion models have shown great potential for PET image reconstruction, where the primary goal is to restore high-quality images from extremely noisy data [38], [39]. The diffusion process involves gradually corrupting full-dose PET images $X'_0$ into Gaussian noisy images $X'_T$ by incrementally adding random noise, which can be formalized as a Markov chain process:

$$q(X'_t \mid X'_{t-1}) = \mathcal{N}(X'_t; \sqrt{1-\beta_t} X'_{t-1}, \beta_t I) \quad (1)$$

where $\beta_t$ represents the variance schedule that controls the amount of noise added. When $T$ is sufficiently large, $X'_T$ approaches a standard Gaussian distribution. Let $\alpha_t = 1 - \beta_t$ and $\bar{\alpha}_t = \prod_{i=1}^{t} \alpha_i$. Given a timestep $t$, $X'_t$ can be derived as:

$$X'_t = \sqrt{\bar{\alpha}_t} X'_0 + \sqrt{1-\bar{\alpha}_t} \epsilon \quad (2)$$

with $\epsilon \sim \mathcal{N}(\mathbf{0}, \mathbf{I})$ being a sample from the Gaussian distribution.

The reconstruction process, aiming to reverse the diffusion process and recover full-dose PET images $X'_0$, is also formulated as a Markov chain:

$$p_\theta(X'_{t-1} \mid X'_t) = \mathcal{N}(X'_{t-1}; \boldsymbol{\mu}_\theta(X'_t, t), \boldsymbol{\Sigma}_\theta(X'_t, t)) \quad (3)$$

This Gaussian transition involves learnable mean $\boldsymbol{\mu}_\theta$ and variance $\boldsymbol{\Sigma}_\theta$. For the DDPM model, $\boldsymbol{\mu}_\theta$ and $\boldsymbol{\Sigma}_\theta$ are given by:

$$\boldsymbol{\mu}_\theta(X'_t, t) = \frac{1}{\sqrt{\alpha_t}} (X'_t - \frac{\beta_t}{\sqrt{1-\bar{\alpha}_t}} \boldsymbol{\epsilon}_\theta(X'_t, t)) \quad (4)$$

$$\boldsymbol{\Sigma}_\theta(X'_t, t) = \frac{1-\bar{\alpha}_{t-1}}{1-\bar{\alpha}_t} \beta_t \quad (5)$$

All parameters except $\boldsymbol{\epsilon}_\theta$ are fixed. Here, $\boldsymbol{\epsilon}_\theta$ represents a learnable component that estimates the noise added to the data $X'_t$. The DDPM performs denoising by predicting the noise that has been added to the data.

## B. Transformer Models

Transformers were originally proposed in the field of natural language processing. Subsequently, vision transformer [17] adapted this architecture to computer vision by splitting an image into a sequence of patches. Unlike convolutional layers which are limited to capturing local features, vision transformers can effectively model long-range dependencies between patches. This capability has enabled vision transformers to achieve good performance in various pattern recognition tasks, such as image classification [40] and object detection [41]. However, for PET reconstruction tasks, both local and global features are equally important. Therefore, a pure vision transformer, while excelling at capturing global context, may not fully leverage local information, which is crucial for tasks requiring precise reconstruction.

## III. METHOD

Large-scale pre-trained diffusion models on PET image datasets have demonstrated remarkable capabilities in generating full-dose images. However, the lack of appropriate guidance mechanisms leads to generated images failing to maintain semantic consistency and fine-grained detail accuracy. To address these challenges, we designed the ENC and NTC modules in **Fig. 1**, which respectively leverage low-dose PET images as constraints to maintain semantic consistency and transform them into feature representations to ensure fine-grained detail accuracy. Notably, NTC enforces the low-rank and sparse properties of these feature representations. These properties enable the feature representations to suppress substantial interfering information inherent in low-dose PET images while retaining rich features at a lower dimensionality.

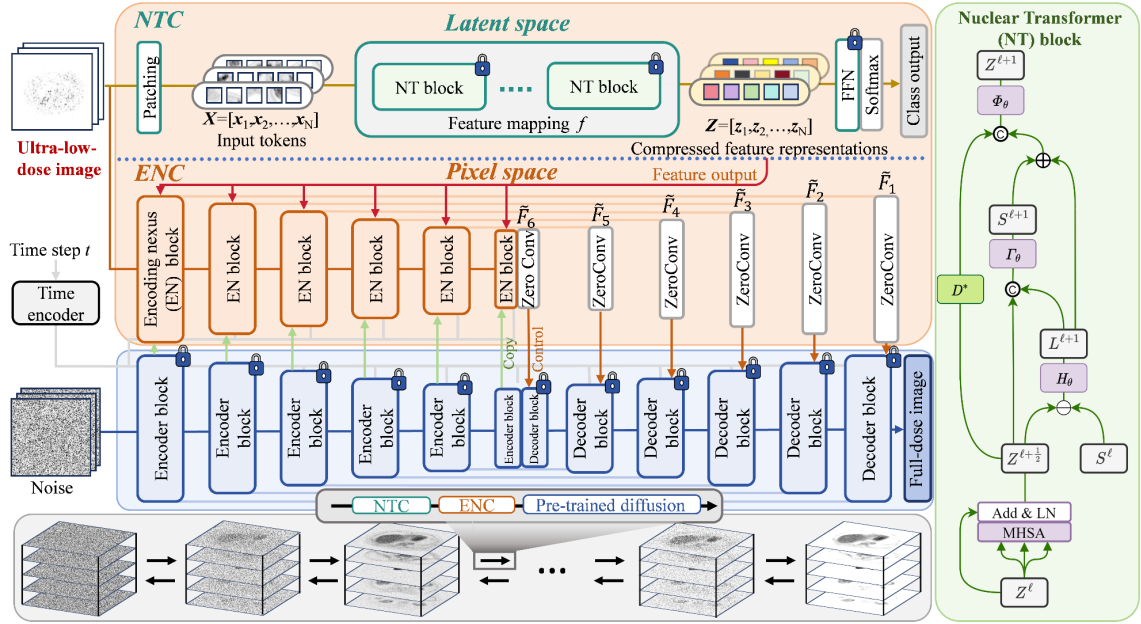

**Fig. 1.** Overview of the DCDM framework. The architecture freezes the parameters of a pre-trained diffusion model and integrates a double-constraint controller into the encoder architecture. The controller comprises a NTC module and an ENC module to enforce hierarchical guidance.

**Fig. 2** displays a comparative 3D visualization evaluating the rank and sparsity of NTC and other recognition models in their compressed feature embeddings, synchronized with the feature extraction progression. The sequence of three sparsity-focused plots illustrates how zero-count distributions evolve across samples for the three models during feature extraction progression, unveiling changing sparsity patterns. The rightmost plot details rank values, quantifying structural differences. Collectively, these visualizations afford insights into how NTC and other recognition models diverge in preserving rank and sparsity during compression, thereby highlighting NTC's superiority in sustaining feature fidelity. This is fundamental to deciphering model behavior in the latent space and informing architecture optimization.

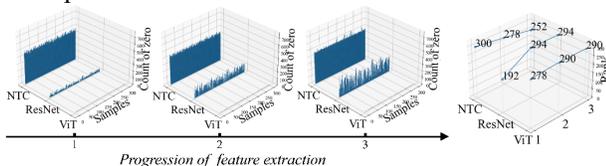

**Fig. 2.** Comparative 3D visualization of sparsity and rank for NTC (Ours) and other recognition model in compressed feature representations.

### A. Adaptive Controller with Double Constraints

**Nuclear Transformer Constraint (NTC):** To extract useful information from ultra-low-dose PET images, we introduce an NTC module that combines long-range dependency modeling with redundancy reduction. This constraint utilizes the advantage of Transformer to capture spatial correlations between non-local tokens while enforcing feature sparsity through a low-rank sparsity regularization scheme. A unified mathematical model is defined below to perform optimization refinement extraction.

Let $X = [x_1, x_2, \cdots, x_B] \in \mathbb{R}^{c \times h \times w \times B}$ denote input ultra-low-dose PET images, where $c, h, w$ and $B$ represent the number of channels, height, width and batch size of input images, respectively. Then, we apply a convolution layer to generate patch embedding $\tilde{X} = [\tilde{x}_1, \tilde{x}_2, \cdots, \tilde{x}_N] \in \mathbb{R}^{D \times N}$. Here, $\tilde{x}_i \in \mathbb{R}^D$ denotes the $i$-th patch embedding, and $N = hwB / p^2$ represents the total number of patch embeddings. $(p, p)$ specify the size of each patch. Each $\tilde{x}_i$ serves as a *token*, directly interchangeable with a patch embedding. Let $\boldsymbol{Z} = f(\tilde{\boldsymbol{X}}) = [z_1, z_2, \cdots, z_N] \in \mathbb{R}^{d \times N}$ be the random variable, with $z_i \in \mathbb{R}^d$ being the token representations. To obtain the input representations, a feature mapping $f: \tilde{\boldsymbol{X}} \in \mathbb{R}^{D \times N} \to \boldsymbol{Z} \in \mathbb{R}^{d \times N}$ is employed to transform the patch embedding $\tilde{\boldsymbol{X}}$, which may have a potentially nonlinear and multi-modal distribution, into a (piecewise) linearized and compact feature representation $\boldsymbol{Z}$.

For the NTC module, its architecture consists of multiple nuclear Transformer blocks (NT blocks), denoted as $L$. The work can be represented as:

$$f: \tilde{\boldsymbol{X}} \xrightarrow{f^1} \boldsymbol{Z}^1 \cdots \to \boldsymbol{Z}^{\ell-1} \xrightarrow{f^\ell} \boldsymbol{Z}^\ell \cdots \to \boldsymbol{Z}^L = \boldsymbol{Z} \quad (6)$$

where $f^\ell: \mathbb{R}^{d \times N} \to \mathbb{R}^{d \times N}$, $1 \leq \ell \leq L$ is the mapping of the $\ell$-th layer. Given $\tilde{\boldsymbol{X}}$ as input, we use $\boldsymbol{Z}^\ell = [z_1^\ell, z_2^\ell, \cdots, z_N^\ell] \in \mathbb{R}^{d \times N}$ to denote the output of the first $\ell$ layers, and $\boldsymbol{U}_{[K]} = (\boldsymbol{U}_k)_{k=1}^K$ to represent the set of bases of all Gaussians such that the $k$-th Gaussian has mean $\mathbf{0} \in \mathbb{R}^d$ and covariance $\boldsymbol{\Sigma}_k \geq \mathbf{0} \in \mathbb{R}^{d \times d}$. To maximize the information gain and low-rank sparse representations for the final token representation, we define the low-rank sparse rate reduction objective function as follows:

$$\max_{f \in F} E_{\boldsymbol{Z}}[\Delta R(\boldsymbol{Z}; \boldsymbol{U}_{[K]}) - \lambda_1 \|\boldsymbol{L}\|_* - \lambda_2 \|\boldsymbol{S}\|_1]$$
$$= E_{\boldsymbol{Z}}[R(\boldsymbol{Z}) - R^c(\boldsymbol{Z}; \boldsymbol{U}_{[K]}) - \lambda_1 \|\boldsymbol{L}\|_* - \lambda_2 \|\boldsymbol{S}\|_1] \quad (7)$$
$$s.t. \quad \boldsymbol{Z} = \boldsymbol{L} + \boldsymbol{S}$$

where the nuclear norm $\|\boldsymbol{L}\|_*$ enforces low-rank property and the sparsity-promoting term $\|\boldsymbol{S}\|_1$ enhances the sparsity of the final representations $\boldsymbol{Z} = f(\boldsymbol{X})$. Here, the objective involves $\lambda_1$ and $\lambda_2$ as positive parameters. $R$ and $R^c$ are lossy coding rates, whose estimates for the number of bits required to encode the sample up to a precision $\varepsilon > 0$ using a Gaussian codebook, both unconditionally (for $R$) and conditioned on the samples being drawn from $\boldsymbol{U}_k$ summed over all $k$ (for $R^c$), and are defined as:

$$R(\boldsymbol{Z}) = \frac{1}{2} \log \det(\boldsymbol{I} + \frac{n}{N\varepsilon^2} \boldsymbol{Z}^T \boldsymbol{Z}), \quad R^c(\boldsymbol{Z}|\boldsymbol{U}_{[K]}) = \sum_{k=1}^K R(\boldsymbol{U}_k^T \boldsymbol{Z}) \quad (8)$$

Next, we transform the object of Eq. (7) into the equivalent form of an unrolled optimization as follows:

$$\min_{\boldsymbol{Z}} E_{\boldsymbol{Z}} R^c(\boldsymbol{Z}; \boldsymbol{U}_{[K]}) \quad (9)$$

$$\min_{\boldsymbol{L},\boldsymbol{S},\boldsymbol{Z}}[\lambda_1 \|\boldsymbol{L}\|_* + \lambda_2 \|\boldsymbol{S}\|_1 - R(\boldsymbol{Z})] \quad s.t. \quad \boldsymbol{Z} = \boldsymbol{L} + \boldsymbol{S} \quad (10)$$

To solve the problem in Eq. (9), minimizing locally by performing a step of gradient descent on $R^c(\boldsymbol{Z}|\boldsymbol{U}_{[K]})$ leads to the multi-head subspace self-attention (MHSA) operation [42]. MSSA is defined as:

$$\text{MHSA}(\boldsymbol{Z}|\boldsymbol{U}_{[K]}) = \frac{p}{N\varepsilon^2}[\boldsymbol{U}_1, \cdots, \boldsymbol{U}_K] \begin{bmatrix} (\boldsymbol{U}_1^*\boldsymbol{Z})\text{softmax}((\boldsymbol{U}_1^*\boldsymbol{Z})^*(\boldsymbol{U}_1^*\boldsymbol{Z})) \\ \vdots \\ (\boldsymbol{U}_K^*\boldsymbol{Z})\text{softmax}((\boldsymbol{U}_K^*\boldsymbol{Z})^*(\boldsymbol{U}_K^*\boldsymbol{Z})) \end{bmatrix} \quad (11)$$

And the subsequent intermediate representation is:

$$\boldsymbol{Z}^{\ell+1/2} = \boldsymbol{Z}^\ell - \eta \nabla_{\boldsymbol{Z}} R^c(\boldsymbol{Z}^\ell | \boldsymbol{U}_{[K]})$$
$$\approx (1 - \eta \cdot \frac{p}{N\varepsilon^2}) \boldsymbol{Z}^\ell + \eta \cdot \frac{p}{N\varepsilon^2} \cdot \text{MHSA}(\boldsymbol{Z}^\ell | \boldsymbol{U}_{[K]}) \quad (12)$$

where $\eta$ is a positive learning rate hyperparameter.

For the optimization function in Eq. (10), the expansion term $R(\boldsymbol{Z})$ promotes diversity and non-collapse of the representation. However, achieving this benefit on large-scale datasets has been challenging due to the poor scalability of the gradient $\nabla_{\boldsymbol{Z}} R(\boldsymbol{Z})$, which involves matrix inversion. To address this issue, we have adopted an alternative method to balance representational diversity and sparsity. Specifically, we introduce an incoherent or orthogonal dictionary $\boldsymbol{D} \in \mathbb{R}^{d \times d}$. Utilizing this dictionary, we aim to achieve a sparser and lower-rank representation $\boldsymbol{Z}^{\ell+1}$. The dictionary $\boldsymbol{D}$ is global, meaning it is used to sparse all tokens simultaneously. Under the assumption $\boldsymbol{D}^*\boldsymbol{D} \approx \boldsymbol{I}_n$, $R(\boldsymbol{Z}^{\ell+1}) \approx R(\boldsymbol{D}\boldsymbol{Z}^{\ell+1}) \approx R(\boldsymbol{Z}^{\ell+1/2})$. Therefore, we approximately solve Eq. (10) using the following optimization:

$$\min_{\boldsymbol{L},\boldsymbol{S},\boldsymbol{Z}}[\lambda_1 \|\boldsymbol{L}\|_* + \lambda_2 \|\boldsymbol{S}\|_1 + \frac{1}{2} \|\boldsymbol{Z}^{\ell+1/2} - \boldsymbol{D}\boldsymbol{Z}\|_2^2] \quad s.t. \quad \boldsymbol{Z} = \boldsymbol{L} + \boldsymbol{S} \quad (13)$$

This minimization task in Eq. (13) can be solved by using the Alternating Direction Method of Multipliers (ADMM), a classical optimization algorithm that operates by decoupling the original complex problem into smaller, computationally tractable sub-problems to streamline the solution process. At the same time, we linearize the function $F(\boldsymbol{Z})$ at the iteration $\hat{\boldsymbol{Z}}^{\ell+1} := \boldsymbol{L}^{\ell+1} + \boldsymbol{S}^{\ell+1}$ to obtain the following subproblems:

$$\begin{cases} \boldsymbol{L}^{\ell+1} = \arg\min_{\boldsymbol{L}} \{\frac{\rho}{2} \|\boldsymbol{L} + \boldsymbol{S}^\ell - \boldsymbol{Z}^{\ell+1/2}\|_2^2 + \lambda_1 \|\boldsymbol{L}\|_*\} \\ \boldsymbol{S}^{\ell+1} = \arg\min_{\boldsymbol{S}} \{\frac{\rho}{2} \|\boldsymbol{L}^{\ell+1} + \boldsymbol{S} - \boldsymbol{Z}^{\ell+1/2}\|_2^2 + \lambda_2 \|\boldsymbol{S}\|_1\} \\ \boldsymbol{Z}^{\ell+1} = \arg\min_{\boldsymbol{Z}} \{\frac{\rho}{2} \|\boldsymbol{L}^{\ell+1} + \boldsymbol{S}^{\ell+1} - \boldsymbol{Z}\|_2^2 + <\nabla F(\boldsymbol{L}^{\ell+1} + \boldsymbol{S}^{\ell+1}), \boldsymbol{Z} > \\ \qquad + \frac{1}{2\eta} \|\boldsymbol{Z} - \boldsymbol{L}^{\ell+1} - \boldsymbol{S}^{\ell+1}\|_2^2\} \end{cases} \quad (14)$$

where $F(\boldsymbol{Z}) := \frac{1}{2} \|\boldsymbol{Z}^{\ell+1/2} - \boldsymbol{D}\boldsymbol{Z}\|_2^2$ and $\eta > 0$. The use of traditional methods for solving Eq. (14) requires hundreds of iterations to achieve satisfactory results. Additionally, the low-rank term $\|\boldsymbol{L}\|_*$ and the sparse term $\|\boldsymbol{S}\|_1$ present significant challenges, and the selection of the hyper-parameters (e.g. $\lambda_1, \lambda_2, \rho, \eta$, etc.) is a very tricky task. However, leveraging deep networks to automatically learn adapters and parameters from sample data has proven to be a simple and effective approach. Therefore, each of the three sub-problems can be considered as a particular instance of the proximal gradient method combined with deep networks. These subproblems can be solved by iterating between the following update steps:

$$\begin{cases} \boldsymbol{L}^{\ell+1} = H_\theta(\boldsymbol{Z}^{\ell+1/2} - \boldsymbol{S}^\ell) \\ \boldsymbol{S}^{\ell+1} = \Gamma_\theta(\boldsymbol{Z}^{\ell+1/2}, \boldsymbol{L}^{\ell+1}) \\ \boldsymbol{Z}^{\ell+1} = \Phi_\theta(\boldsymbol{L}^{\ell+1} + \boldsymbol{S}^{\ell+1}, \boldsymbol{D}^*\boldsymbol{Z}^{\ell+1/2}) \end{cases} \quad (15)$$

where the operators $H_\theta$, $\Gamma_\theta$ and $\Phi_\theta$ are learned by the fully connected feed-forward network (FFN). Specifically, the FNN consists of two linear layers and a SiLU activation function, mathematically formulated as:

$$\text{FFN}(\boldsymbol{Y}) = \text{SiLU}(\boldsymbol{W}_1 \times \boldsymbol{Y} + \boldsymbol{b}_1) \times \boldsymbol{W}_2 + \boldsymbol{b}_2 \quad (16)$$

where $Y$ is an input feature with dimension $d$. $W_1, W_2, b_1$ and $b_2$ represent the weights and biases of the two linear layers, respectively. In the FFN projection process, the dimension of $Y$ is compressed to $d/4$ after passing the first linear layer, and then restored to $d$ via the second linear layer. It is worth noting that when the input of the FFN contains two variables, we concatenate them into a single variable before input. To utilize the feature extraction capabilities from the image classification task, a classification layer composed of a FFN and a softmax activation function is incorporated following the feature mapping $f$ during the training process. This layer is designed to map $Z$ to a class output. Additionally, cross-entropy loss is employed as the loss function for training this classification layer. The weights of the NTC are denoted as $\Theta_N$.

*Encoding Nexus Constraint (ENC):* To combine the feature encodings of NTC and further exploit information from ultra-low-dose PET images, we design an ENC module that includes six pairs of encoding nexus blocks (EN blocks) and zero convolution (ZeroConv) layers. The detailed structure of the proposed EN block is depicted in **Fig. 3**.

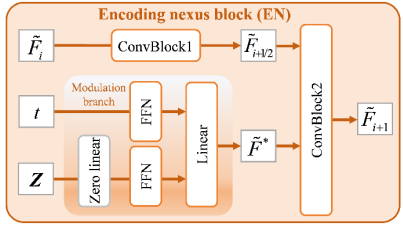

**Fig. 3.** The detailed structure of the encoding nexus (EN) block in the ENC module.

Given the time step $t$ from the pre-trained diffusion model (details provided in **Section III. B**) and the compressed feature representation $Z$ from NTC, these two inputs are separately processed by a FFN and a FFN equipped with a zero linear layer. The zero linear, in which both the weight matrix and bias are initialized to zero, is introduced to enhance the training stability of the ENC module. Subsequently, they are fused into the modulation feature $\tilde{F}^*$ via addition and linear projection. Let the output of each EN block be $\tilde{F} = \{\tilde{F}_1, \tilde{F}_2, \cdots, \tilde{F}_6\}$. For the $i$-th EN Block, $\tilde{F}_i$ is first transformed into $\tilde{F}_{i+1/2}$ by the first convolution block (ConvBlock1). Then the modulation feature $\tilde{F}^*$ provides guidance to $\tilde{F}_{i+1/2}$ by channel-wise addition. Last, the output feature $\tilde{F}_{i+1}$ is generated by the second convolution block (ConvBlock2). The EN Block process is summarized as:

$$\tilde{F}^* = \text{Linear}(\text{Zero linear}(\text{FFN}(Z)+\text{FFN}(t))) \quad (17)$$

$$\tilde{F}_i = \text{ConvBlock2}(\text{ConvBlock1}(\tilde{F}_{i-1})+\tilde{F}^*) \quad (18)$$

Furthermore, we employ six ZeroConv layers to establish a connection between $\tilde{F}$ and the pre-trained diffusion model. The ZeroConv, a convolution layer with weights and biases initialized to zero, is designed to help mitigate random noise gradients from the pre-trained diffusion model.

### B. Overview of DCDM

*Pre-training of the Diffusion Model:* The pre-training of the diffusion model is a foundational step that enables DCDM to effectively learn from full-dose PET images and adapt to subsequent fine-tuning with double constraints. During this phase, the model is trained as a reconstruction network to generate high-quality full-dose PET images by learning the underlying data distribution.

In this process, the model progressively adds Gaussian noise into full-dose PET images $X' \in \mathbb{R}^{c \times h \times w \times B}$ over the time step $t$. A neural network $U$, composed of an encoder $U_E$ and a decoder $U_D$, is used to estimate and remove added noise at each time step through Markov chain. The encoder $U_E$ contains six encoder blocks, each identical to the EN block but without the modulation branch. These blocks are connected via a convolutional layer with a kernel size of 3 and a stride of 2. The decoder $U_D$ consists of six decoder blocks, each structured similarly to the encoder blocks but using nearest-neighbor interpolation (scale factor 2) to restore the original resolution. The weights of the encoder and decoder are denoted as $\Theta_E$ and $\Theta_D$, respectively. Mathematically, the pre-training process is described by:

$$X'_t = \gamma_t X' + \sigma_t \epsilon \quad (19)$$

$$\hat{\epsilon} = U_D(U_E(X'_t, t; \Theta_E), t; \Theta_D) \quad (20)$$

where $\gamma_t = \sqrt{\bar{\alpha}_t}$ and $\sigma_t = \sqrt{1-\bar{\alpha}_t}$ denote the scale factors of the full-dose PET images and of the noise. $\epsilon \sim \mathcal{N}(0, I)$ denotes the added noise, and $\hat{\epsilon}$ is the predicted noise. The training objective is to minimize the loss function:

$$\mathcal{L}_{rec} = \|\hat{\epsilon} - \epsilon\|_2^2 \quad (21)$$

where $\|\cdot\|_2^2$ represents the $L_2$ norm.

*Double Constraints for Pre-trained Diffusion:* To achieve high-quality ultra-low-dose PET image reconstruction, the proposed DCDM method incorporates a double-constraint controller into a pre-trained diffusion model, with the overall framework shown in **Fig. 1**. By enhancing flexibility in handling varying dose levels and optimizing reconstruction accuracy using prior information, DCDM effectively reconstructs images that closely full-dose PET images under different dose conditions.

Before training the DCDM, we lock these trained parameters $\Theta_N$, $\Theta_E$, and $\Theta_D$. The encoder $\Theta_E$ of the pre-trained diffusion model, composed of 6 encoder blocks, shares the same architecture as the EN block without the modulation feature $\tilde{F}^*$ branch. Thus, ENC initializes its parameters $\Theta_C$ by copying $\Theta_E$ from $U_E$.

During training, DCDM is trained using pairs of full-dose PET images $X'$ and ultra-low-dose PET images $X$. The low-dose PET images $X$ are transformed into a compressed feature representation $Z$ via NTC, serving as the first constraint. Meanwhile, ENC encodes $X$ into a control signal $\tilde{F}$ conditioned on $Z$ and time step $t$. Subsequently, $X'$ with random noise is fed into $U$, and the double constraints are injected into intermediate blocks of decoder $U_D$ via ZeroConv layers. The training process is summarized as follows:

$$Z = \text{NTC}(X; \Theta_N) \quad (22)$$

$$\tilde{F} = \text{ENC}(X, Z, t; \Theta_C) \quad (23)$$

$$\hat{\epsilon} = U_D(U_E(X'_t, t; \Theta_E), \text{ZeroConv}(\tilde{F}), t; \Theta_D) \quad (24)$$

For PET image reconstruction at different dose levels, the only trainable parameter is $\Theta_C$, which significantly reduces training resource requirements and enables $U$ to generate the corresponding full-dose PET image. A detailed comparison of parameters is provided in **Section IV. D**. In summary, the DCDM algorithm is provided for ultra-low-dose PET reconstruction, as seen in **Algorithm 1**.

---

**Algorithm 1: DCDM**

**Prior learning**

1: **Input:** $X$, $X'$, $\Theta_E$, $\Theta_D$, and $\Theta_N$
2: Sample $X'_t$ via Eq. (19)
3: $\tilde{F} = \text{ENC}(X, \text{NTC}(X;\Theta_N), t; \Theta_C)$
4: $\hat{\epsilon} = U_D(U_E(X'_t, t; \Theta_E), \text{ZeroConv}(\tilde{F}), t; \Theta_D)$
5: **Optimization:** $\|\hat{\epsilon} - \epsilon\|_2^2$
6: **Output:** Learned $\Theta_C$

**Iterative reconstruction**

1: **Input:** $X$, $T$, $\alpha$, and $\beta$
2: $X'_T \sim \mathcal{N}(\mathbf{0}, \mathbf{I})$
3: $Z = \text{NTC}(X;\Theta_N)$
4: **For** $t = T-1$ **to** $0$ **do**
5:    $\tilde{F} = \text{ENC}(X, Z, t; \Theta_C)$
6:    $\hat{\epsilon} = U_D(U_E(X'_t, t; \Theta_E), \text{ZeroConv}(\tilde{F}), t; \Theta_D)$
7:    $\mathbf{z} \sim \mathcal{N}(\mathbf{0}, \mathbf{I})$ if $t > 1$, else $\mathbf{z} = \mathbf{0}$
8:    $X'_{t-1} = \frac{1}{\sqrt{\alpha_t}}(X'_t - \frac{\beta_t}{\sqrt{1-\bar{\alpha}_t}}\hat{\epsilon}) + \sqrt{\frac{1-\bar{\alpha}_{t-1}}{1-\bar{\alpha}_t}\beta_t}\mathbf{z}$
9: **End for**
10: **Output:** $X'_0$

---

### C. Adaptive DCDM for an Unknown DRF Scenario

While traditional deep learning-based models have shown remarkable success in reconstructing high-quality images from low-quality inputs, most neural networks trained on specific low-dose levels struggle to effectively reconstruct images at other low-dose levels. In clinical practice, the DRF of low-dose PET images is often unknown, which limits the robustness of traditional deep learning-based reconstruction models when handling an unknown DRF scenario.

To address this issue, we propose the DCDM framework for unknown DRF reconstruction. As depicted in **Fig. 4(b)**, this framework integrates five distinct ENC modules, each trained on low-dose PET images with DRF values of 100, 50, 20, 10, and 4, respectively, alongside an NTC module and a pre-trained diffusion model. For a PET image with unknown DRF, the NTC module first projects it into a class output. Subsequently, the most compatible ENC module is selected to generate a control signal, which is used to guide the pre-trained diffusion model for reconstruction. NTC enables decomposition of the complex unknown dose distribution into multiple single dose distributions and process each simplified distribution by a separate ENC. Compared with the traditional deep learning-based model that directly processes the unknown dose distribution, the DCDM demonstrates adaptability and flexibility. A comprehensive comparison of unknown DRF reconstruction is provided in **Section IV. B**.

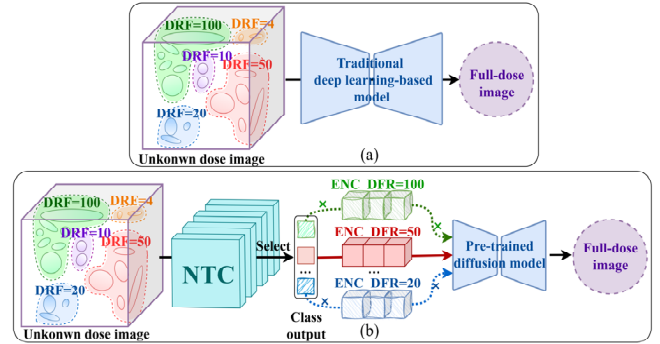

**Fig. 4.** Comparative illustration of a traditional deep learning-based model and the DCDM framework for unknown DRF reconstruction.

## IV. EXPERIMENTS

### A. Experimental Setup

In this section, the performance of DCDM is compared with state-of-the-arts at varying dose levels, including U-Net [43], MPRNet [14], ViT [17], Pix2Pix [44], IDDPM [30] and ControlNet [45]. ViT-Rec and ViT-Cls denote the application of the ViT architecture for PET reconstruction and classification tasks, respectively. To ensure comparability and fairness of the experiments, all methods are conducted on the same datasets. Open-source code is available at: https://github.com/yqx7150/DCDM.

*Datasets:* Two different datasets were used in the experiments. *UDPET* is a public dataset from the MICCAI 2024 ultra-low-dose PET imaging challenge, including low-dose and full-dose images with DRFs of 4, 10, 20, 50, and 100. Low-dose images were created by subsampling full scans and were perfectly aligned with corresponding full-dose images. Each patient provided 673 axial 2D slices, cropped to 256×256 by removing background for training and testing. Data were acquired using the uEXPLORER whole-body PET system for $^{18}$F-FDG imaging. Each DRF level contains data from 101 patients, totaling 67,973 2D slices for training. Additionally, 1,346 extra slices were chosen to assess the validity across different dose levels. *Clinical* dataset was selected from 10 patients to test generalization. For each patient, imaging results were provided consisting of 450 axial 2D image slices of 250×250. These slices were zero-padded to 256×256 for testing. Data were sampled at unknown DRF from the DigitMI 930 PET/CT scanner (RAYSOLUTION Healthcare Co., Ltd.), featuring all-digital PET detectors with an axial field-of-view (AFOV) of 30.6 cm. Each scan covered 4 to 8 beds, with scan times ranging from 45 seconds to 3 minutes per bed. Low-dose PET data were obtained by resampling list-mode data into random intervals, retaining the random data per cycle and discarding the remaining data.

*Parameter Configuration:* The maximum timestep $T$ was set to 1,000. The diffusion model was pre-trained on full-dose images for 300,000 optimization iterations at a resolution of 256×256 pixels with a batch size of 6. NTC was trained on multiple low-dose images for 100,000 optimization iterations. ENC was trained on each specific low-dose level for 100,000 optimization iterations. Throughout all training phases, the AdamW optimizer was employed with a learning rate of $1\times10^{-4}$. The training and testing experiments were conducted using 2 NVIDIA GeForce RTX 3090 GPUs with 24 GB memory each.

*Performance Evaluation:* To quantitatively measure the error caused by DCDM, the peak signal-to-noise ratio (PSNR), structural similarity (SSIM), Fréchet inception distance (FID) and learned perceptual image patch similarity (LPIPS) [46] are used to evaluate the quality of reconstruction images. To further assess the effectiveness on the *Clinical* dataset, additional clinical metrics are employed. These include the difference of the maximum standardized uptake value ($\Delta SUV_{max}$) and the difference of the mean standardized uptake value ($\Delta SUV_{mean}$), both calculated between the reference and reconstructed images for the lesion, as well as the signal-to-noise ratio (SNR), coefficient of variation (CoV), and contrast ratio (CR). The specific expressions for SNR, CoV, and CR are as follows:

$$\text{SNR} = SUV_{mean\_lesion} / SD_{liver} \tag{25}$$

$$\text{CoV} = SD_{liver} / SUV_{mean\_liver} \tag{26}$$

$$\text{CR} = SUV_{max\_lesion} / SUV_{mean\_liver} \tag{27}$$

where $SUV_{mean\_lesion}$ and $SUV_{mean\_liver}$ represent the mean standardized uptake value for the lesion and liver, respectively. $SUV_{max\_lesion}$ is the maximum standardized uptake value for the lesion, and $SD_{liver}$ is the standard deviation of the liver.

### B. Reconstruction Experiments

*Comparison of UDPET Public Dataset with Known Dose Levels:* To verify the advantages of the proposed DCDM, we conduct a comparative experiment with different methods. **Table I** illustrates a quantitative comparison of various state-of-the-art methods for ultra-low-dose PET reconstruction across different DRFs. The results show that the proposed DCDM outperforms existing methods like U-Net, MPRNet, ViT-Rec, Pix2Pix, IDDPM, and ControlNet in terms of PSNR, SSIM, FID, and LPIPS. For example, when the DRF is set to 100, DCDM attains the highest PSNR value of 40.12 dB and SSIM value of 0.9725, while also achieving the lowest FID of 21.40 and LPIPS of 0.0356. Similarly, DCDM attains the highest PSNR of 47.15 dB, SSIM of 0.9905, and the lowest FID of 15.94 and LPIPS of 0.0200 at a DRF value of 4. The results indicate that the double-constraint controller in DCDM effectively enhances the performance of the diffusion model, enabling it to produce higher quality PET images with better structural similarity and perceptual quality.

**Fig. 5** presents a visual comparison of the reconstruction results on the *UDPET* public dataset across different known dose levels. The proposed DCDM shows significant advantages over other state-of-the-art methods. In terms of overall image quality, the PET images reconstructed by DCDM display higher clarity and preserve more fine details of anatomical structures. For example, the image reconstructed by DCDM closely resembles the full-dose image in terms of intensity distribution and structural sharpness at a DRF value of 100. In the region of interest (ROI), marked by the red box, DCDM demonstrates greater accuracy in reconstructing the details of lesions or areas of interest. The lesion edges are clearer, and the internal intensity distribution is more reasonable compared to the full-dose image. In contrast, other methods such as U-Net, MPRNet, and ViT-Rec appear somewhat blurry in certain anatomical structures within the ROI. The error maps below each reconstructed PET image visually represent the differences between the reconstructed images and the full-dose images. DCDM exhibits lower error magnitudes across the entire image and a more uniform error distribution without significant localized high-error regions. Conversely, other methods show some regions with relatively concentrated errors, indicated by the red arrows. These regions may correspond to areas with complex anatomical structures or significant intensity changes. Overall, the visual comparison highlights the effectiveness of the double-constraint controller in DCDM in improving the performance of the diffusion model and enabling the production of higher quality PET images with better structural similarity and perceptual quality.

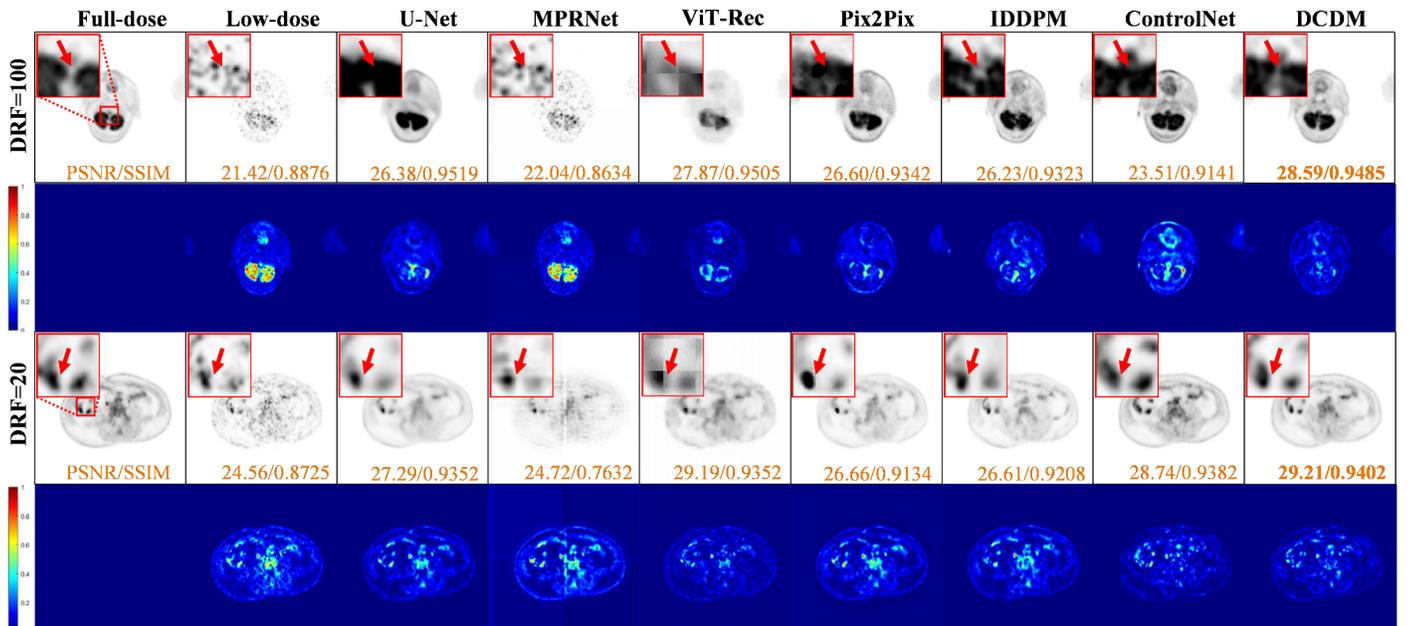

**Fig. 5.** Reconstruction results on the *UDPET* public dataset at different known dose levels. From left to right: Full-dose, low-dose, and reconstructions by U-Net, MPRNet, ViT-Rec, Pix2Pix, IDDPM, ControlNet and DCDM (Ours).

TABLE I
COMPARISON OF STATE-OF-THE-ART METHODS IN TERMS OF AVERAGE PSNR, SSIM, FID, AND LPIPS UNDER VARIOUS DRFS ON THE *UDPET* DATASET. ↓ REPRESENTS THE SMALLER THE BETTER, AND ↑ REPRESENTS THE BIGGER THE BETTER. THE **BOLD** AND *ITALIC* FONTS INDICATE THE OPTIMAL AND SUB-OPTIMAL VALUES, RESPECTIVELY.

| DRF | Metric | U-Net | MPRNet | ViT-Rec | Pix2Pix | IDDPM | ControlNet | DCDM |
|---|---|---|---|---|---|---|---|---|
| 100 | PSNR(dB) ↑ | 26.61 | 24.22 | 25.55 | 37.98 | 39.57 | 39.73 | **40.12** |
|  | SSIM ↑ | 0.9565 | 0.9012 | 0.9392 | 0.9489 | 0.9685 | 0.9718 | **0.9725** |
|  | FID ↓ | 63.38 | 181.04 | 117.70 | 31.10 | 32.68 | 22.92 | **21.40** |
|  | LPIPS ↓ | 0.0514 | 0.0999 | 0.0913 | 0.0655 | 0.0478 | 0.0390 | **0.0356** |
| 50 | PSNR(dB) ↑ | 27.74 | 28.54 | 27.07 | 38.64 | 39.88 | 40.45 | **41.25** |
|  | SSIM ↑ | 0.9634 | 0.9300 | 0.8306 | 0.9574 | 0.9729 | 0.9772 | **0.9779** |
|  | FID ↓ | 52.59 | 159.88 | 130.09 | 39.65 | 26.79 | 22.85 | **21.84** |
|  | LPIPS ↓ | 0.0449 | 0.0798 | 0.2000 | 0.0665 | 0.0397 | 0.0400 | **0.0348** |
| 20 | PSNR(dB) ↑ | 31.37 | 29.02 | 29.77 | 40.73 | 41.42 | 42.60 | **42.67** |
|  | SSIM ↑ | 0.9727 | 0.8813 | 0.9011 | 0.9702 | 0.9802 | 0.9815 | **0.9831** |
|  | FID ↓ | 39.82 | 193.14 | 112.26 | 35.46 | 30.13 | **20.56** | *22.02* |
|  | LPIPS ↓ | 0.0368 | 0.1885 | 0.1573 | 0.0551 | 0.0371 | 0.0300 | **0.0300** |
| 10 | PSNR(dB) ↑ | 32.77 | 34.09 | 32.01 | 41.98 | 43.26 | 43.47 | **44.15** |
|  | SSIM ↑ | 0.9775 | 0.9803 | 0.9050 | 0.9745 | 0.9805 | 0.9850 | **0.9862** |
|  | FID ↓ | 33.05 | 48.84 | 102.30 | 40.26 | 30.88 | 20.63 | **20.05** |
|  | LPIPS ↓ | 0.0304 | **0.0292** | 0.1206 | 0.0500 | 0.0300 | 0.0300 | *0.0300* |
| 4 | PSNR(dB) ↑ | 36.17 | 38.24 | 32.58 | 44.69 | 46.75 | 46.48 | **47.15** |
|  | SSIM ↑ | 0.9830 | 0.9896 | 0.8910 | 0.9848 | 0.9853 | 0.9900 | **0.9905** |
|  | FID ↓ | 22.68 | 24.02 | 94.39 | 33.95 | 25.76 | 17.38 | **15.94** |
|  | LPIPS ↓ | 0.0222 | **0.0129** | 0.1367 | 0.0424 | 0.0300 | 0.0200 | *0.0200* |

*Comparison of Clinical Dataset with an Unknown DRF Scenario:* To evaluate the performance of different methods on the *Clinical* dataset with an unknown DRF, **Fig. 6** presents a visual comparison of the reconstruction results. The three rows present the full-dose reference image, the unknown DRF input, and the reconstructions from U-Net, MPRNet, ViT-Rec, Pix2Pix, IDDPM, ControlNet, and DCDM. Visually, DCDM's reconstruction closely resembles the full-dose image. It displays sharper anatomical details and reduced noise, particularly in the region of interest highlighted by the red box. Quantitative assessment using PSNR and SSIM metrics further underscores DCDM's superiority. DCDM achieves the highest PSNR of 33.10 dB and SSIM of 0.9495, significantly outperforming other methods such as U-Net, which attains a PSNR of 26.33 dB and SSIM of 0.9403, and IDDPM, with a PSNR of 28.34 dB and SSIM of 0.9355. The error maps displayed below each reconstruction further demonstrate DCDM's effectiveness in minimizing residual noise and suppressing artifacts. These results indicate that DCDM possesses robust generalization capabilities for an unknown DRF scenario.

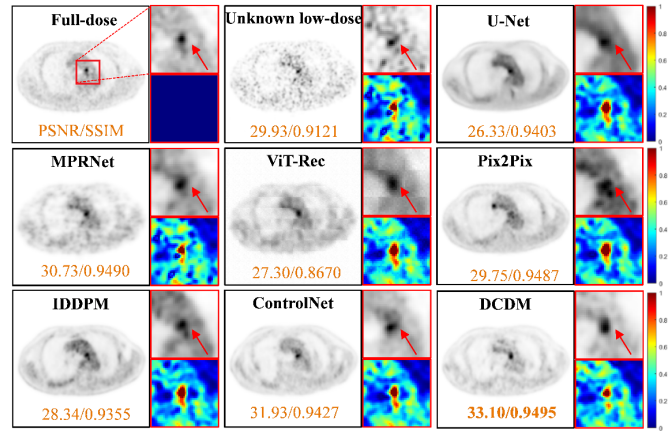

**Fig. 6.** Reconstruction results on the *Clinical* dataset at an unknown DRF. The three rows present Full-dose, unknown low-dose, and reconstructions by U-Net, MPRNet, ViT-Rec, Pix2Pix, IDDPM, ControlNet, and DCDM (Ours).

TABLE II
PERFORMANCE COMPARISON OF STATE-OF-THE-ART METHODS IN TERMS OF THE AVERAGE $\Delta SUV_{MAX}$, $\Delta SUV_{MEAN}$(*E-3), SNR, COV, AND CR UNDER UNKNOWN DRF ON THE *CLINICAL* DATASET.

| Clinical dataset | $\Delta SUV_{max}$↓ | $\Delta SUV_{mean}$↓ | SNR↑ | CoV↓ | CR↑ |
|---|---|---|---|---|---|
| U-Net | 1.21 | 0.64 | 3.68 | 0.1011 | 0.4913 |
| MPRNet | 1.11 | 0.58 | 1.90 | 0.2497 | 0.6683 |
| ViT-Rec | 1.23 | 0.66 | 2.13 | 0.2052 | 0.5918 |
| Pix2Pix | 1.27 | 0.70 | 2.08 | 0.1756 | 0.4913 |
| IDDPM | 1.61 | 0.63 | 1.90 | 0.2174 | 0.5881 |
| ControlNet | 1.03 | 0.57 | 6.85 | 0.0669 | 0.7147 |
| **DCDM** | **0.91** | **0.49** | **8.46** | **0.0601** | **0.8081** |

**Table II** assesses the performance of state-of-the-art methods on the *Clinical* dataset for PET images with unknown DRF. The evaluation focuses on several clinical metrics, including the average $\Delta SUV_{max}$, $\Delta SUV_{mean}$, SNR, CoV, and CR. Smaller values of $\Delta SUV_{max}$ and $\Delta SUV_{mean}$ indicate closer alignment with reference metabolism. Higher SNR values denote better signal dominance over noise. Lower CoV values signify more homogeneous tissue intensity. Higher CR values reflect stronger lesion-to-liver contrast. The results show that DCDM demonstrates superior performance across all these metrics. It achieves the lowest $\Delta SUV_{max}$ of 0.91 and $\Delta SUV_{mean}$ of 0.49, outperforming ControlNet, which records values of 1.03 and 0.57, and IDDPM, with values of 1.61 and 0.63. This indicates that DCDM preserves metabolic activity more precisely. Notably, DCDM also attains the highest SNR of 8.46 and CR of 0.8081, along with the lowest CoV of 0.0601. These results highlight DCDM's effectiveness in noise suppression, tissue uniformity, and lesion visibility. Overall, the findings show that DCDM's double-constraint framework, which integrates nuclear norm regularization for low-rank feature extraction and adaptive encoding for diffusion guidance, mitigates the degradation of unknown DRF and ensures reliable clinical reconstruction.

### C. Ablation Study

To evaluate the impact of the NTC and ENC modules in DCDM, an ablation study on the *UDPET* dataset at a DRF value of 100 is presented in **Table III**. The "Ultra-low-dose" scenario exhibits severe degradation, with an average PSNR of 20.84 dB and an average SSIM of 0.8489. Eliminating both constraints leads to a performance boost, achieving an average PSNR of 37.49 dB and an average SSIM of 0.9545. This highlights the baseline capability of the diffusion model when conditioned on the low-dose PET image. However, the FID remains at 48.01 and the LPIPS at 0.0472, suggesting structural and perceptual differences. Introducing only the Encoding Nexus Constraint slightly elevates the SSIM to 0.9584 and decreases the FID to 33.30, indicating its role in controlling diffusion. The full configuration of DCDM, with both constraints applied, yields the optimal results such as a PSNR value of 38.24 dB, an SSIM value of 0.9607, an FID value of 33.11, and an LPIPS value of 0.0344. This implies that the NTC's low-rank feature extraction and long-range dependency, along with the ENC's encoding modeling, contributes to refining compressed representations, thus enhancing quantitative metrics and perceptual quality.

TABLE III
COMPARISON IN TERMS OF AVERAGE PSNR, SSIM, FID, AND LPIPS UNDER A DRF VALUE OF 100 ON THE *UDPET* DATASET.

| DRF=100 | PSNR ↑ | SSIM ↑ | FID ↓ | LPIPS ↓ |
|---|---|---|---|---|
| Ultra-low-dose | 20.84 | 0.8489 | 246.77 | 0.1426 |
| (w/o) NTC&ENC | 37.49 | 0.9545 | 48.01 | 0.0472 |
| (w/o) NTC | 37.56 | 0.9584 | 33.30 | 0.0387 |
| DCDM | **38.24** | **0.9607** | **33.11** | **0.0344** |

## V. DISCUSSION

We have demonstrated that spatially constraining the diffusion process effectively improves reconstruction stability and convergence speed. For the proposed DCDM, the ability of NTC to effectively capture and utilize relevant features from low-dose PET images is crucial for enhancing the accuracy and reliability of reconstructed images. The following analysis examines the compressed feature representations generated by NTC and their impact on reconstruction quality.

*Analysis of Compressed Feature Representations:* **Fig. 7** presents t-SNE visualizations of compressed feature representations extracted by ViT-Cls, ResNet, and the proposed NTC, illustrating their ability to distinguish ultra-low-dose PET images across different DRFs. The NTC-generated features exhibit tightly clustered groups for each DRF, with clear separation between dose levels, indicating strong discriminative power. In contrast, ResNet produces scattered, overlapping clusters, particularly between high-DRF classes such as DRF values of 50 and 100, reflecting its inability to capture dose-specific structural dependencies due to limited global context modeling. ViT-Cls shows moderate clustering but with inter-class overlaps such as DRF values of 10 and 20, suggesting that while Transformer-based self-attention aids in global feature extraction, the absence of nuclear norm regularization leads to residual redundancy in low-dose representations.

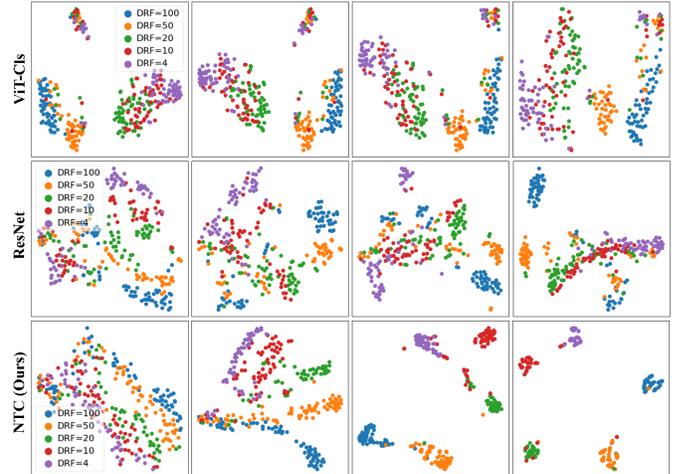

**Fig. 7.** t-SNE visualizations of compressed feature representation process across different methods such as ViT-Cls, ResNet and NTC (Ours).

## VI. CONCLUSIONS

In this study, we presented a novel framework incorporating a nuclear Transformer for ultra-low-dose PET reconstruction. The framework, with its double-constraint controller consisting of NTC and ENC modules, offers a solution to the challenge of maintaining image quality while significantly reducing radiation exposure. The NTC module captures long-range dependencies and maintains low-rank features, while the ENC module precisely controls the pre-trained diffusion model. Our experiments on the *UDPET* public dataset and the *Clinical* dataset with unknown DRF demonstrated that the proposed model not only outperforms state-of-the-art methods in known dose scenarios but also shows strong generalization ability in unknown DRF scenarios. Future research directions may include further refinement of the model architecture, exploration of larger and more diverse datasets, and integration of multi-modal imaging data to further enhance reconstruction performance and clinical applicability.